\pdfoutput=1

\documentclass[11pt]{article}

\usepackage{acl}

\usepackage{times}
\usepackage{latexsym}

\usepackage[T1]{fontenc}

\usepackage[utf8]{inputenc}

\usepackage{microtype}

\usepackage{inconsolata}

\usepackage{lipsum}
\usepackage{enumitem}
\usepackage{hyperref}
\usepackage{todonotes}
\usepackage{multirow}
\usepackage{booktabs}
\usepackage{subcaption}
\usepackage{graphicx}
\usepackage{float}

\definecolor{orange}{HTML}{E66100}
\definecolor{purple}{HTML}{5D3A9B}

\title{Unpacking Tokenization: Evaluating Text Compression and its Correlation with Model Performance}

\newcommand{\authorspace}{\hspace{9pt}}

\author{Omer Goldman$^{\beta\gamma}$ \authorspace Avi Caciularu$^{\gamma}$ \authorspace Matan Eyal$^{\gamma}$ \\
\bf Kris Cao$^{\delta}$\authorspace Idan Szpektor$^{\gamma}$\authorspace Reut Tsarfaty$^{\gamma}$\\\\ {$^\beta$Bar-Ilan University}\authorspace{$^\gamma$Google Research}\authorspace{$^\delta$Google DeepMind}\\
\footnotesize{\texttt{\{ogoldman,avica,matane,kriscao,szpektor,reutt\}@google.com}}
}

\begin{document}
\maketitle
\begin{abstract}
Despite it being the cornerstone of BPE, the most common tokenization algorithm, the importance of compression in the tokenization process is still unclear.
In this paper, we argue for the theoretical importance of compression, that can be viewed as $0$-gram language modeling where equal probability is assigned to all tokens.%
We also demonstrate the empirical importance of compression for downstream success of pre-trained language models. %
We control the compression ability of several BPE tokenizers by varying the amount of documents available during their training: %
from 1 million documents to a character-based tokenizer equivalent to no training data at all. We then pre-train English language models based on those tokenizers and fine-tune them over several tasks. 
We show that there is a correlation between tokenizers' compression and models' downstream performance, suggesting that compression is a reliable intrinsic indicator of tokenization quality. 
These correlations are more pronounced for generation tasks (over classification) %
or for smaller models (over large ones). 
We replicated a representative part of our experiments on Turkish and found similar results, confirming that our results hold for languages with typological characteristics dissimilar to English.
We conclude that building better compressing tokenizers is a fruitful avenue for further research and for improving overall model performance.
\end{abstract}

\section{Introduction}

\begin{figure*}[t]
    \centering
    \begin{subfigure}[b]{0.49\linewidth}
        \includegraphics[width=\linewidth]{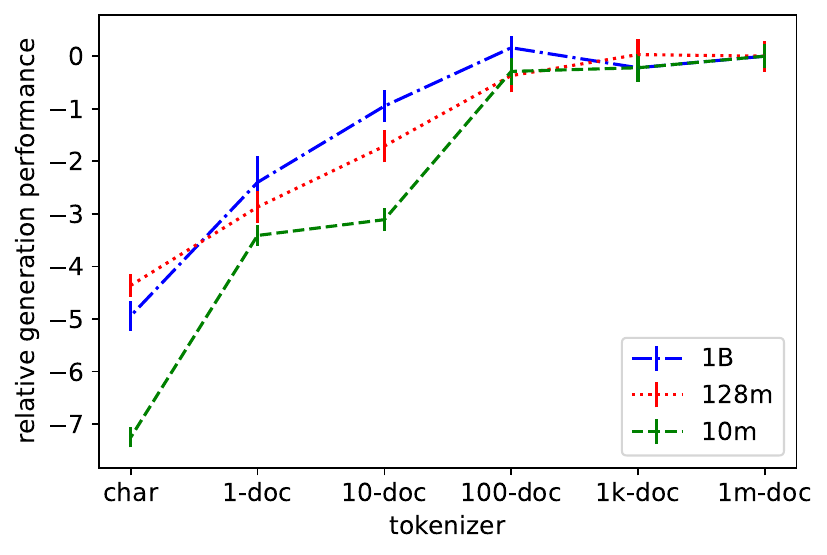}
        \caption{English}
        \label{fig:English_relative_generation}
    \end{subfigure}
    \begin{subfigure}[b]{0.49\linewidth}
        \includegraphics[width=\linewidth]{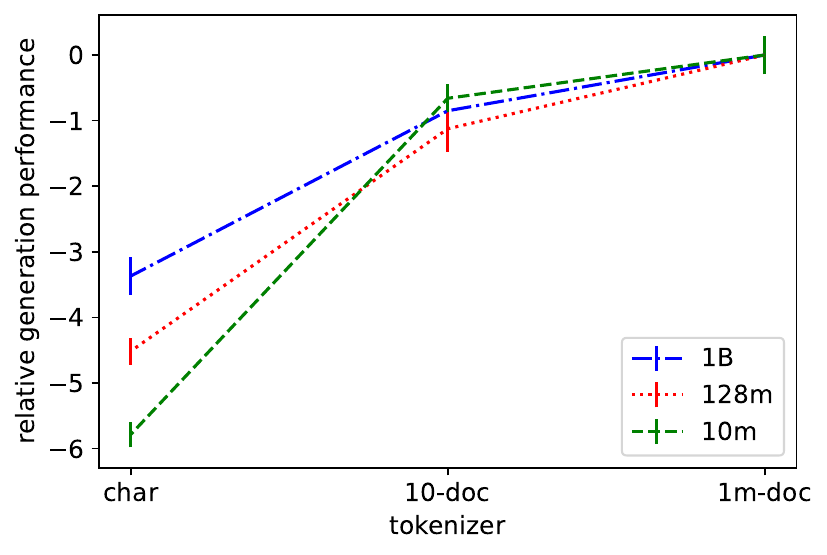}
        \caption{Turkish}
        \label{fig:turkish_relative_generation}
    \end{subfigure}
    \caption{Generation performance of the various models averaged over both generation tasks. For each model size the results are presented as relative compared to the \textsc{1m-doc} model.}
    \label{fig:relative_generation}
\end{figure*}

While language modeling pipelines employ a multitude of sophisticated techniques to achieve success in many NLP tasks, their presupposed tokenization, i.e., the step of discretizing text into processable units, is often done with less scrutiny or deviation from the common practices. %
This tokenization stage, which segments space-delimited words into subwords, forms the foundation of most large language models (LLMs; \citealp{touvron2023llama, geminiteam2023gemini, groeneveld2024olmo}, inter alia) and influences their modus operandi in subsequent usage. 
Among other open questions regarding tokenization, it is unclear whether tokenization is even needed \cite{clark-etal-2022-canine, xue-etal-2022-byt5, keren2022breaking} and how much poor tokenization influences model performance, especially for non-English languages \cite{klein-tsarfaty-2020-getting,rust-etal-2021-good,gueta2023large}.%

As the tokenizers serve language models, it is straightforward that the primary method to assess their quality is by measuring their contribution to the model performance over the NLP tasks it is meant to solve, i.e., evaluating the tokenizers on tasks extrinsic to tokenization itself. However, this method requires pretraing expensive LLMs whenever an evaluation of a tokenizer is needed. For this reason an intrisic indicator of tokenization quality is warrented. And indeed the literature is teeming with intrinsic evaluations of tokenization.
For example,  \citet{sennrich-etal-2016-neural} used text compression as the main indicator of the tokenizer's intrinsic quality, whereas \citet{bostrom-durrett-2020-byte} suggested assessing tokenizers based on segmentation overlap with a morphologically segmented reference.

In this paper we carefully distinguish between intrinsic and extrinsic evaluation of tokenizers,
and examine to what extent they are correlated. %
As a specific intrinsic evaluation we focus on compression, the metric underpinning BPE \cite{sennrich-etal-2016-neural},
the most prevalent tokenization algorithm that requires character co-occurrence statistics over a large corpus of raw text to achieve minimal length in tokens. %
We control the tokenizer's ability to compress by limiting its \textit{support}, i.e., the amount of data available in the tokenizer's training corpus. By doing so we skew the statistics available to the tokenizer.%

We compared tokenizers trained with a million supporting documents to ones trained on less and less data, down to a single document, and to a character-level tokenizer, equivalent to {\em zero support}. We then pre-trained from scratch copies of a decoder-only transformer-based model \cite{vaswani2017attention}, with the different tokenizers, and finetuned them on several downstream tasks. In this work we hypothesize that the downstream success should be correlated with the compression ability of the underlying tokenizers. We experimented with three model sizes, tokenizers of six different volumes of supporting data, and two languages, English and Turkish.

\begin{figure*}[t]
    \centering
    \begin{tabular}{r c} \toprule
    \textsc{\textbf{Tokenizer}} & \textsc{\textbf{Sentence}} \\
    \midrule
        \textsc{char} & \texttt{\_T h i s \_i s \_a b o u t \_c o m p r e s s i n g \_t o k e n i z e r s}\\  
        \textsc{1-doc} & \texttt{\_Th is \_is \_a b ou t \_comp re ss ing \_to k en i z ers} \\
        \textsc{10-doc} & \texttt{\_This \_is \_about \_comp res sing \_to k en iz ers} \\
        \textsc{100-doc} & \texttt{\_This \_is \_about \_comp ress ing \_tok en izers} \\
        \textsc{1k-doc} & \texttt{\_This \_is \_about \_comp ressing \_to ken izers} \\
        \textsc{1m-doc} & \texttt{\_This \_is \_about \_compress ing \_token izers} 
        \\ \bottomrule
    \end{tabular}
    \caption{Six tokenizers differing in the amount of supporting documents tokenizing the same sentence. Note that better compression is achieved with more support.}
    \label{fig:compression}
\end{figure*}

Our results show that in terms of intrinsic performance, the tokenizers' compression ability %
is highly influenced by the amount of supporting data, with tokenizers trained on a minimal amount of data %
having tokenized texts more than 60\% longer compared to the best compressing tokenizer. However, the discrepancy in compression is significantly more marked for less frequent words. %

Extrinsically, we also 
found that downstream success monotonically increases with the increase in the tokenizer's support.
The correlation between the intrinsic and extrinsic measures of tokenization quality points to the conclusion that better compressing tokenizers is a desired goal on the road to better language models. A conclusion that may be true even for models dealing with other modalities \cite{NEURIPS2021_6a30e32e,ronen2023vision}.

While we evaluated the downstream performance on both classification and generative tasks, we observed that the correlation to compression is stronger for the latter type of tasks. %
This discrepancy could be attributed to the fact that generative tasks require the use of the tokenizer more extensively than in classification tasks, aligning with the number of generation steps involved. We therefore conclude that tokenization's effect is better assessed through generation tasks, rather than classification tasks. %

Our results also show that smaller models are especially vulnerable to poor tokenizations, with the smallest $10m$ parameter model suffering from more significant drops in performance compared to our largest $1B$ model.  %
Finally, experimentation with Turkish revealed the same trends, %
ruling out the option of an English-specific phenomenon.%

In the remainder of the paper we will describe the common practices in assessing tokenizers (\autoref{sec:related}) and argue for the theoretical sensibility of compression as an intrinsic tokenization evaluation (\autoref{sec:theory}). We will then describe our experiments (\autoref{sec:setup}) and their results (\autoref{sec:results}).

\section{Measuring Tokenization Quality}
\label{sec:related} 

From the very early days of NLP, models have always assumed text discretized to tokens as input \cite{winograd-1971-procedures}. For the most part, these tokens were whitespace %
separated words, but in the recent decade non-trivial %
tokenization algorithms have surfaced \cite{mikolov2012subword, sennrich-etal-2016-neural,ataman2018evaluation}, primarily to deal with unseen tokens without smoothing techniques or other convoluted methods \cite{chen1999empirical}. The underlying reasoning behind all modern tokenization methods is that some subwords, e.g., morphemes, may carry independent information that is of value to the model even if the word as a whole is rare or unseen. Therefore, better tokenization is assumed to improve models' performance over rare words, while also carrying computational benefits, like smaller models and the elimination of unknown tokens. %

It is not surprising then, that whenever tokenizers %
are presented or tested, they are usually accompanied with an array of evaluations 
that assess 
the tokenization's influence on the model's downstream success, mostly on translation tasks \cite{kudo-2018-subword, provilkov-etal-2020-bpe, vilar-federico-2021-statistical, saleva-lignos-2023-changes}, although monolingual tasks are also used \cite{yehezkel-pinter-2023-incorporating}.

Other works circle back and assess tokenization with respect to the desiderata it is supposed to serve as a stand-alone algorithm, independently from the model trained on top of it. This is done usually in addition to evaluation over downstream performance. 
However, most works disagree on the desiderata themselves. Many emphasize alignment to linguistically meaningful units \cite{klein-tsarfaty-2020-getting, hofmann-etal-2021-superbizarre, hofmann-etal-2022-embarrassingly, gow-smith-etal-2022-improving} or to human cognitive preferences in general \cite{beinborn-pinter-2023-analyzing}.\footnote{This line of works may view tokenization as a continuation of unsupervised morphemic segmentation \cite{creutz-lagus-2002-unsupervised, virpioja2013morfessor}.} Others include analyses of token length and frequency \cite{bostrom-durrett-2020-byte, yehezkel-pinter-2023-incorporating}, mostly in addition to the above, assuming that ideal tokenizers use longer and more frequent tokens. %

The two types of tokenization evaluations, extrinsic over downstream success and intrinsic over a plethora of metrics, are usually not compared directly. They are only used to demonstrate the superiority of a specific tokenizer, and the relations between the \textit{evaluation approaches} is glossed over. In this work we explicitly focus on compression as a potential intrinsic indicator of tokenization quality, as has been suggested in past works in other settings \cite{galle-2019-investigating,gutierrez2023languages}, and check to what extent it is correlated with extrinsic downstream success. We conclude that compression is a desideratum for tokenization not only due to its theoretical virtue, expanded on in the next section, but first and foremost because it correlates with downstream performance.

\section{The Role of Compression in Tokenization}
\label{sec:theory}

In the realm of intrinsic measures for evaluating tokenization quality, compression particularly stands out in prominence. It has garnered considerable attention, notably due to its pivotal role as the cornerstone of the byte pair encoding tokenization algorithm \citep[BPE;][]{schuster2012japanese,sennrich-etal-2016-neural}, an algorithm that initially conceived for general data compression purposes \cite{gage1994new}.

Given data composed of sequences of atomic symbols, the algorithm minimizes the overall data length, for a given dictionary budget, by iteratively substituting a new symbol in place of the symbol pair most frequently occurring in a large corpus of \textit{supporting} data.
In the domain of language modeling, the symbols are usually characters and the supporting corpus is a subset of the text designated to be used as a training set for the language model which the tokenizer is meant to serve.

But in a sense, compression-driven tokenization may be viewed as language modeling in and of itself. 
Consider that language models are aimed at assessing and possibly maximizing the likelihood of produced texts expressed as a product of token probabilities,
$$P(\mathbf{x})=\prod_k P(x_k|\mathbf{x}_{1:k-1}),$$
where $x_k$ is the $k$th token in a sentence $\mathbf{x}$. Compression limits the lower bound on this product of fractions by minimizing the number of operands, i.e., minimizing the length of the sequence.

In terms of $n$-gram language modeling, where the probability of each token is approximated as dependent only on a context of length $n-1$,
$$P(\mathbf{x})\approx\prod_k P(x_k|\mathbf{x}_{k-(n-1):k-1}),$$
and the probability of $x_k$ given its context is further approximated by the number of appearances of the relevant $n$-gram in a training corpus,
$$P(x_k|\mathbf{x}_{k-(n-1):k-1}) \propto N(\mathbf{x}_{k-(n-1):k}),$$
A compressor may be considered a $0$-gram language model, where 
the relevant $n$-gram is of length $0$ and 
the probability of each token is not even a function of its own frequency in the training data, setting uniformly,
$$P(x_i) = |V|^{-1},$$
where $|V|$ is the vocabulary size. 

Although simplistic when thinking about language modeling with predefined whitespace-separated words, this type of objective is sensible when considering that it is used to determine the symbols themselves.

From this point of view, prioritizing compression as an indicator for tokenization quality is very reasonable. Since BPE optimizes an approximation, albeit crude, of the downstream objective, doing better under this approximated objective should translate into better downstream performance which will justify the focus on compression as a metric.

Moreover, from an information theoretic perspective, Shannon's source coding theorem \cite{shannon1948mathematical} links the limit on the compression to the entropy of the source of the data to be compressed. As language models aim to increase the log-likelihood of texts, hence decrease the entropy of the distribution, they inadvertently also increase the possible compression of the texts. Our claim is that this relationship is symmetric, and BPE tokenizers, as they compress texts, may also inadvertently increase their log-likelihood.

We set to empirically examine our hypothesis by assessing the correlation between the tokenizer's compression ability and the performance of language models of various sizes over a set of downstream tasks. 

To explicitly control the compression ability, while fixing any other intervening factors as much as possible, we deal only with BPE tokenizers. This is in contrast with other works that compared compression across different tokenization algorithms \cite{galle-2019-investigating,schmidt2024tokenization}. 
To create BPE tokenizers with varied compression rate 
we recall that BPE's maximal compression is guaranteed only over its supporting corpus. %
Normally, for large enough corpora, a minimal discrepancy is assumed between the character distribution in the corpus and the ``true'' distribution in the language. In this work however, %
we explicitly emphasize and expand this discrepancy by limiting the size of the support to a great extent. We will show that this intervention severely hinders the compression capacity of the tokenizer and that it also leads to deteriorating downstream performance.

\section{Experimental Setup}
\label{sec:setup}

\subsection{English Experiments}
\paragraph{Tokenizers} We trained six different tokenizers with dictionary size of up to 32$k$ tokens.\footnote{For some tokenizers with little supporting data, there were less than 32$k$ strings and sub-strings so the vocabulary in practice was smaller. See \autoref{sec:appendix} for details.} Each tokenizer was supported by a different amount of documents from the model's train set: a million (\textsc{1m-doc}), a thousand (\textsc{1k-doc}), a hundred (\textsc{100-doc}), ten (\textsc{10-doc}), one document (\textsc{1-doc}) and no documents at all (\textsc{char}).
The tokenizers are initialized with all the relevant symbols - the characters of the alphabet, punctuation marks, and all foreign characters that appear on the respective documents. 

\paragraph{Models} For every tokenizer, %
we trained decoder-only transformer-based language models in three sizes, in terms of number of  parameters: \textit{1B}, \textit{128m} and \textit{10m}. The model sizes exclude the parameters dedicated to the embedding layer, as its size may differ across tokenizers. 
See \autoref{sec:appendix} for further details.

\paragraph{Data} Pretraining of the English models was executed %
monolingually using the train split of C4 \cite{2020t5}.

\paragraph{Tasks} To evaluate the tokenizers' contribution to downstream success we finetuned the models over four tasks. Two classification tasks: %
\begin{itemize}[nosep]
    \item \textit{QQP} (Quora Question Pairs\footnote{\url{https://quoradata.quora.com/First-Quora-Dataset-Release-Question-Pairs}}) where the task is to classify 2 questions as duplicates or not.
    \item \textit{MultiNLI} \cite{williams-etal-2018-broad} where the model is tested on natural language inference (NLI) examples from a domain which differs from the ones appearing at the training set.
\end{itemize}
And two generation tasks:
\begin{itemize}[nosep]
    \item \textit{X-Sum} \cite{narayan-etal-2018-dont} where news articles should be summarized to one single sentence.
    \item \textit{QG-QA} (Question Generation over SQuAD; \citealp{rajpurkar-etal-2016-squad}) where the task is to generate questions based on a context paragraph and an answer.
\end{itemize}

\begin{table}[t]
    \centering
    \begin{tabular}{c c c}
    \toprule
        Tokenizer & Token Length & Relative Length \\
        \midrule
        \textsc{1m-doc} & 9,336,052 & --- \\
        \textsc{1k-doc} & 9,541,368 & +2\% \\
        \textsc{100-doc} & 10,489,029 & +12\% \\
        \textsc{10-doc} & 15,126,769 & +62\% \\
        \textsc{1-doc} & 20,647,861 & +121\% \\
        \textsc{char} & 39,480,577 & +323\% \\
        \bottomrule
    \end{tabular}
    \caption{Compression ability of the different tokenizer, accumulative over the development sets of all downstream tasks. Relative length is in comparison to the \textsc{1m-doc} tokenizer.}
    \label{tab:compression}
\end{table}

\subsection{Turkish Experiments} 

To make sure that our results are not due to English-specific phenomena, we repeat a representative subest of the experiments with Turkish, an agglutinative language with higher morphemes-per-word ratio.
for the purpose of intrinsic evaluation, we trained  six Turkish tokenizers, as we did for English. However for extrinsic evaluation, due to the expensive pretraining and finetuning, we trained models only for three tokenizers: \textsc{1m-doc}, \textsc{10-doc}, and \textsc{char}. %
The models were pretrained on the train split of the Turkish part of mC4 \cite{xue-etal-2021-mt5}, and finetuned over three tasks: one classification task,  XNLI \cite{conneau-etal-2018-xnli}, and two generation tasks,  XL-Sum \cite{hasan-etal-2021-xl} and Question Generation over the TQuAD dataset.\footnote{\url{https://tquad.github.io/turkish-nlp-qa-dataset/}}

\section{Results} %
\label{sec:results}

\subsection{Intrinsic Evaluation}

To illustrate the effect of limiting the tokenization support on the compression ability, we measured the accumulative length in tokens of the development sets of all English downstream tasks.

The results, depicted in \autoref{tab:compression}, %
show that providing less support severely impedes the tokenizer ability to compress unseen texts. Note that the inflation in texts' length is not linear. Reducing the supporting data amount by three orders of magnitude, from \textsc{1m-doc} to \textsc{1k-doc}, results in only slightly longer texts, while a reduction in another three orders of magnitude to the \textsc{1-doc} tokenizer carries an effect much more significant.

\begin{table*}[h!]
    \centering
\begin{tabular}{c c c c c}
\toprule
\multirow{2}{*}{\textsc{Tokenizer}} & \multicolumn{4}{c}{\textsc{Task}} \\ & \textsc{QQP} (F1) & \textsc{MultiNLI} (Acc.) & \textsc{XSum} (RougeL) & \textsc{QG-QA} (RougeL) \\ \midrule
\multicolumn{5}{c}{1B params} \\ \midrule
\textsc{1m-doc} & 88.02{\small $\pm$0.18} & 88.24{\small $\pm$0.10} & 47.71{\small $\pm$0.02} & 33.09{\small $\pm$0.41} \\
\textsc{1k-doc} & 87.38{\small $\pm$0.05} & 88.32{\small $\pm$0.07} & 47.53{\small $\pm$0.04} & 32.95{\small $\pm$0.42} \\
\textsc{100-doc} & 88.30{\small $\pm$0.09} & 88.75{\small $\pm$0.11} & 47.69{\small $\pm$0.05} & 32.99{\small $\pm$0.46} \\
\textsc{10-doc} & 87.44{\small $\pm$0.07} & 88.27{\small $\pm$0.09} & 47.07{\small $\pm$0.06} & 30.51{\small $\pm$0.61} \\
\textsc{1-doc} & 86.07{\small $\pm$0.20} & 86.67{\small $\pm$0.25} & 46.33{\small $\pm$0.11} & 28.42{\small $\pm$0.99} \\
\textsc{char} & 83.13{\small $\pm$0.23} & 84.59{\small $\pm$0.65} & 44.69{\small $\pm$0.08} & 24.91{\small $\pm$0.55} \\
\midrule
\multicolumn{5}{c}{128m params} \\ \midrule
\textsc{1m-doc} & 82.13{\small $\pm$0.14} & 85.33{\small $\pm$0.16} & 45.68{\small $\pm$0.04} & 27.66{\small $\pm$0.59} \\
\textsc{1k-doc} & 82.29{\small $\pm$0.06} & 85.45{\small $\pm$0.18} & 45.83{\small $\pm$0.08} & 27.37{\small $\pm$0.59} \\
\textsc{100-doc} & 81.75{\small $\pm$0.28} & 85.07{\small $\pm$0.13} & 45.53{\small $\pm$0.02} & 26.97{\small $\pm$0.64} \\
\textsc{10-doc} & 80.14{\small $\pm$0.19} & 83.76{\small $\pm$0.06} & 45.08{\small $\pm$0.04} & 24.99{\small $\pm$0.62} \\
\textsc{1-doc} & 78.71{\small $\pm$0.19} & 82.30{\small $\pm$0.31} & 44.43{\small $\pm$0.06} & 23.9{\small $\pm$0.60} \\
\textsc{char} & 76.27{\small $\pm$0.23} & 82.10{\small $\pm$0.26} & 43.19{\small $\pm$0.06} & 21.81{\small $\pm$0.43} \\
\midrule
\multicolumn{5}{c}{10m params} \\ \midrule
\textsc{1m-doc} & 71.65{\small $\pm$0.81} & 78.62{\small $\pm$0.22} & 40.92{\small $\pm$0.06} & 22.43{\small $\pm$0.44} \\
\textsc{1k-doc} & 69.97{\small $\pm$0.11} & 79.94{\small $\pm$0.15} & 40.69{\small $\pm$0.08} & 22.13{\small $\pm$0.53} \\
\textsc{100-doc} & 71.26{\small $\pm$0.28} & 78.57{\small $\pm$0.17} & 41.05{\small $\pm$0.02} & 21.59{\small $\pm$0.52} \\
\textsc{10-doc} & 66.51{\small $\pm$0.22} & 75.95{\small $\pm$0.24} & 39.00{\small $\pm$0.04} & 19.73{\small $\pm$0.43} \\
\textsc{1-doc} & 66.25{\small $\pm$0.11} & 78.11{\small $\pm$0.12} & 37.01{\small $\pm$0.08} & 18.61{\small $\pm$0.40} \\
\textsc{char} & 64.01{\small $\pm$0.14} & 75.81{\small $\pm$0.77} & 27.86{\small $\pm$0.04} & 16.92{\small $\pm$0.38} \\
\bottomrule
\end{tabular}
    \caption{Results over all downstream tasks, each in terms of its respective metric. Results are averaged over 5 finetues.}
    \label{tab:main_results}

\bigskip

\begin{minipage}{\columnwidth}

    \centering
    \resizebox{0.95\columnwidth}{!}{
\begin{tabular}{c c c c c}
\toprule
\multirow{2}{*}{\textsc{Model Size}} & \multicolumn{4}{c}{\textsc{Task}} \\ & \textsc{QQP} & \textsc{MultiNLI} & \textsc{XSum} & \textsc{QG-QA} \\ \midrule
1B & 0.714 & 0.600 & 0.943** & 0.943** \\
128m & 0.943** & 0.943** & 0.943** & 1.000** \\
10m & 0.943** & 0.886* & 0.829* & 1.000** \\
\bottomrule
\end{tabular}
}
    \caption{Spearman's $\rho$ coefficient for rank correlation between downstream performance and the tokenizers' support for all model sizes and tasks. Asterisks are used to denote statistically significant correlation.\hspace{\textwidth}* $p<0.05$, ** $p<0.01$}
    \label{tab:spearman}

\end{minipage}\hfill %
\begin{minipage}{\columnwidth}

    \centering
    \resizebox{\columnwidth}{!}{
\begin{tabular}{c c c c c}
\toprule
\multirow{2}{*}{\textsc{Model Size}} & \multicolumn{4}{c}{\textsc{Task}} \\ & \textsc{QQP} & \textsc{MultiNLI} & \textsc{XSum} & \textsc{QG-QA} \\ \midrule
1B & -0.980** & -0.974** & -0.994** & -0.976** \\
128m & -0.971** & -0.863* & -0.988** & -0.949** \\
10m & -0.870* & -0.710 & -0.996** & -0.933** \\
\bottomrule
\end{tabular}
}
    \caption{Pearson's correlation coefficient between downstream performance and the tokenizers' compression as expressed by the dev sets length (see \autoref{tab:compression}),
    for all model sizes and tasks. Asterisks are used to denote statistically significant correlation.\hspace{\textwidth}* $p<0.05$, ** $p<0.01$}
    \label{tab:pearson}
\end{minipage}

\end{table*}

\subsection{Extrinsic Evaluation}

\begin{table}[t]
    \centering
\begin{tabular}{c c c c c}
\toprule
\multirow{2}{*}{\textsc{Tokenizer}} & \multicolumn{3}{c}{\textsc{Task}} \\ & \textsc{XNLI} & \textsc{XLSum} & \textsc{QG-QA} \\ \midrule
\multicolumn{4}{c}{1B params} \\ \midrule
\textsc{1m-doc} & 79.56 & 46.48 & 29.78 \\
\textsc{10-doc} & 79.28 & 45.92 & 29.02 \\
\textsc{char} & 75.18 & 40.00 & 24.69 \\
\midrule
\multicolumn{4}{c}{128m params} \\ \midrule
\textsc{1m-doc} & 76.23 & 45.07 & 27.45 \\
\textsc{10-doc} & 74.83 & 43.40 & 26.87 \\
\textsc{char} & 68.88 & 40.33 & 23.16 \\
\midrule
\multicolumn{4}{c}{10m params} \\ \midrule
\textsc{1m-doc} & 63.60 & 37.16 & 21.76 \\
\textsc{10-doc} & 64.06 & 37.27 & 19.95 \\
\textsc{char} & 63.73 & 35.56 & 16.62 \\
\bottomrule
\end{tabular}
    \caption{Results for the Turkish models over all downstream tasks, each in terms of its respective metric: accuracy for XNLI, and RougeL for XL-Sum and QG-QA.}
    \label{tab:turkish_results}
\end{table}

\autoref{tab:main_results} summarizes the downstream evaluation results for all models and all tokenizers over all four English tasks. Unsurprisingly, it shows that larger models, in terms of parameters, fare better on all tasks. Additionally, it shows that all models perform better on the classification tasks compared to the generation tasks.
Nevertheless, over most tasks and model sizes, there is a clear improvement in performance when the models are equipped with better supported tokenizers. 

Similarly to the intrinsic metric above, the downstream improvement is not linear as well. 
The improvements achieved by updating the tokenizer from the \textsc{1-doc} to the \textsc{10-doc} are more substantial than those from \textsc{1k-doc} to \textsc{1m-doc}, despite the introduction of significantly fewer documents.

The findings for the Turkish models in \autoref{tab:turkish_results} demonstrate analogous patterns, indicating that the results decline as the tokenizer's support diminishes. This is again particularly noticeable in the case of generation tasks.

\subsection{Intrinsic-Extrinsic Correlation}

To assess the correlation between the tokenizer's support and the model's task performance we computed the Spearman's $\rho$ coefficient, %
separately for each task and each model size. {This correlation coefficient was chosen since it refers to the relative rank of each data point, thus it does not ascribe linear importance to the  differences in the absolute number of supporting documents.} %

The results are shown in \autoref{tab:spearman}. Note that due to the small sample size the correlation is statistically significant ($\alpha=0.05$) only for coefficients larger than 0.829. The results show that, for the most part, the tokenizer's support is well correlated with the model's overall success, with the clear exception of classification tasks on the 1B model. %

Even starker correlation appears when measuring the Pearson's correlation coefficient between downstream performance and the compression itself, i.e., to the overall length of the development sets in tokens, from \autoref{tab:compression}. As can be seen in \autoref{tab:pearson}, the inverse correlation between length in tokens and performance is high, with the exception of classification tasks on the 10m model. Note that here as well, the small sample size causes the correlation to be statistically significant only when above 0.729 in absolute value. No numerical correlation could be computed for the Turkish results due to the small sample size.

These results point to generation tasks as better downstream evaluators of tokenizers, as tokenization is less crucial to the models' success over classification tasks.

In addition to assessing the correlation's significance, \autoref{fig:relative_generation} visualizes the effect size  %
for both languages. We averaged the performance over the generation tasks, as the correlation was less significant for classification. The graph depicts, separately for each model size, the performance with the various tokenizers compared to the best supported tokenizer. Notably, while compression consistently correlates with generation performance across all model sizes, the impact is particularly pronounced for smaller models.

The parallels drawn between tokenization and language modeling in \autoref{sec:theory} may provide some explanation to the smaller effect of poorer tokenization on larger models. As we claim that compression is simple language modeling on its own, it is possible that LLMs that are more powerful language models in general are able to allocate resources to compensate for less compressing tokenizers. %

\begin{figure}[t]
    \centering
    \includegraphics[width=\linewidth]{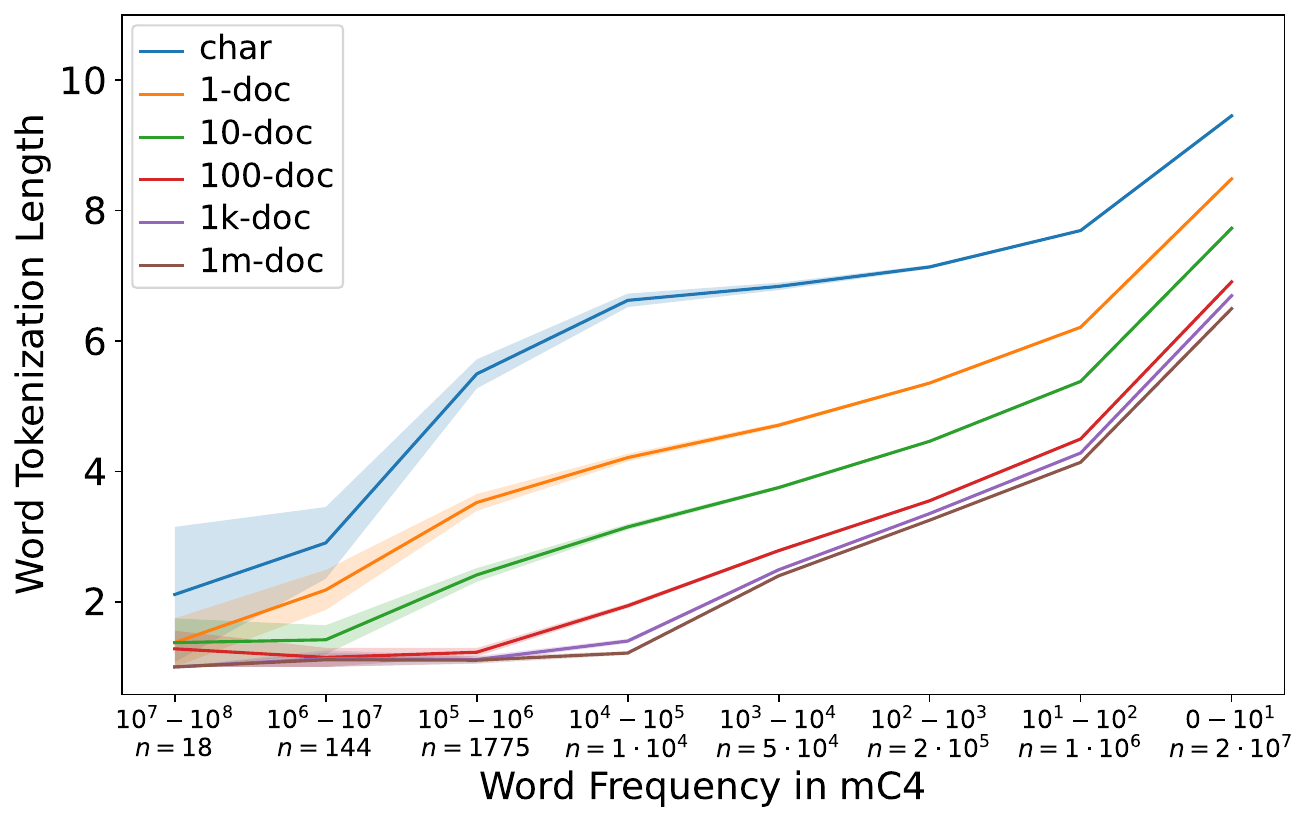}
    \caption{Number of subwords per \textit{English} word as a function of its abundance in 3 million unseen documents. Averaged over orders of magnitude. The number words included in each bin is indicated under the x axis.}
    \label{fig:tokens_vs_freq}
\end{figure}

\begin{figure}[t]
    \centering
    \includegraphics[width=\linewidth]{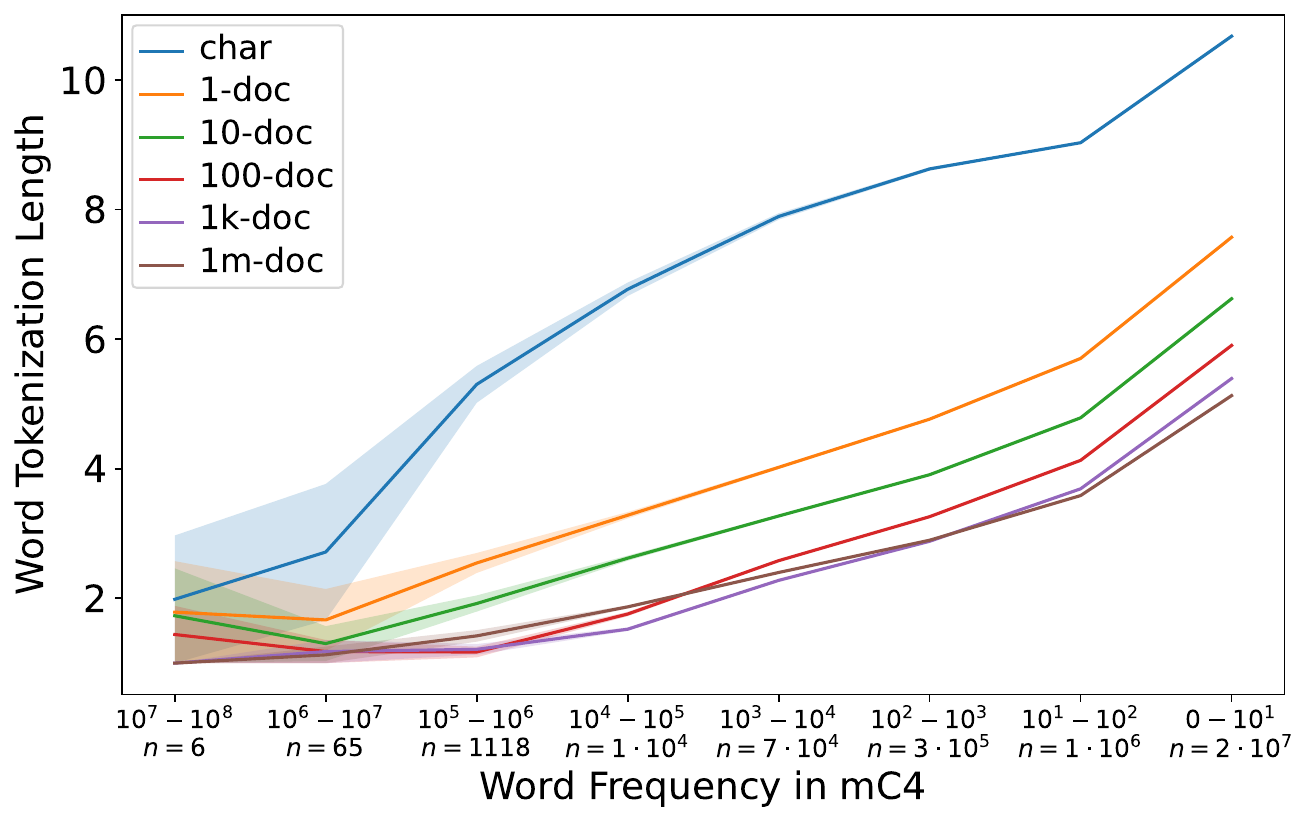}
    \caption{Number of subwords per \textit{Turkish} word as a function of its abundance in 3 million unseen documents. Averaged over orders of magnitude. The number words included in each bin is indicated under the x axis.}
    \label{fig:tokens_vs_freq_tur}
\end{figure}

\begin{figure*}[t]
  \centering
  \begin{subfigure}[b]{0.32\linewidth}
    \includegraphics[width=\linewidth]{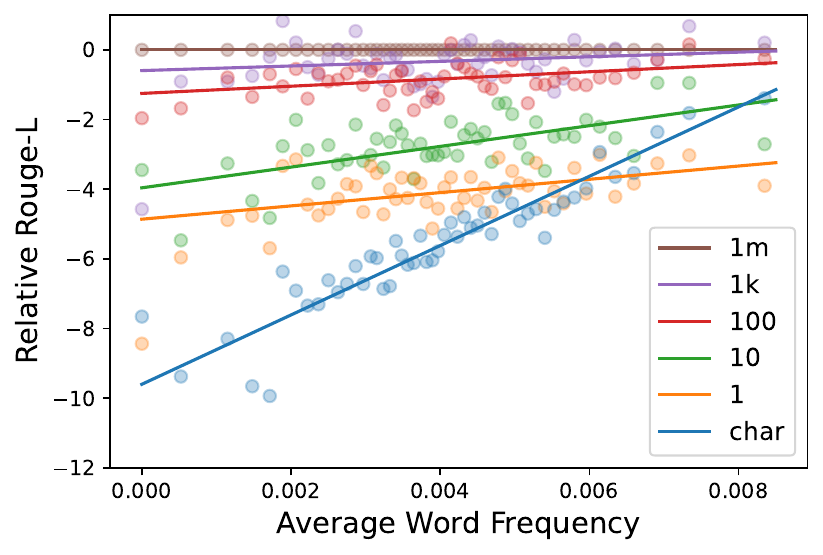}
    \caption{10m params model, XSum}
  \end{subfigure}
  \begin{subfigure}[b]{0.32\linewidth}
    \includegraphics[width=\linewidth]{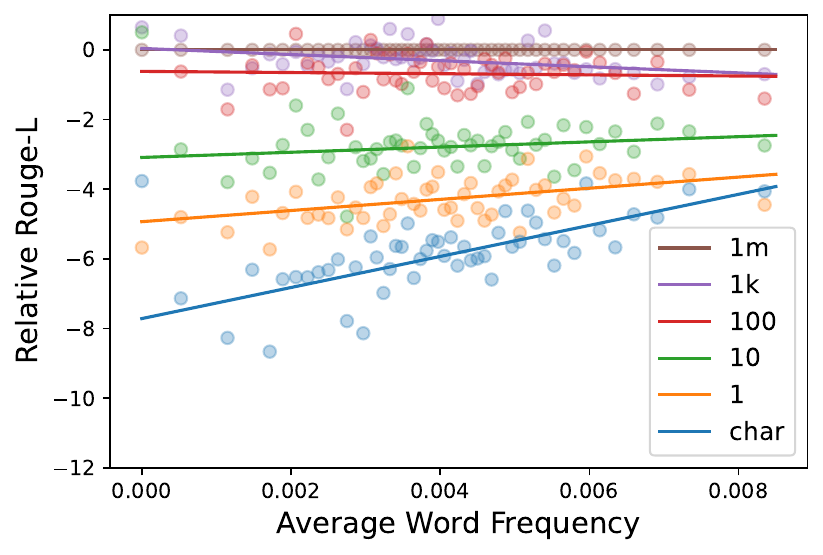}
    \caption{128m params model, XSum}
  \end{subfigure}
  \begin{subfigure}[b]{0.32\linewidth}
    \includegraphics[width=\linewidth]{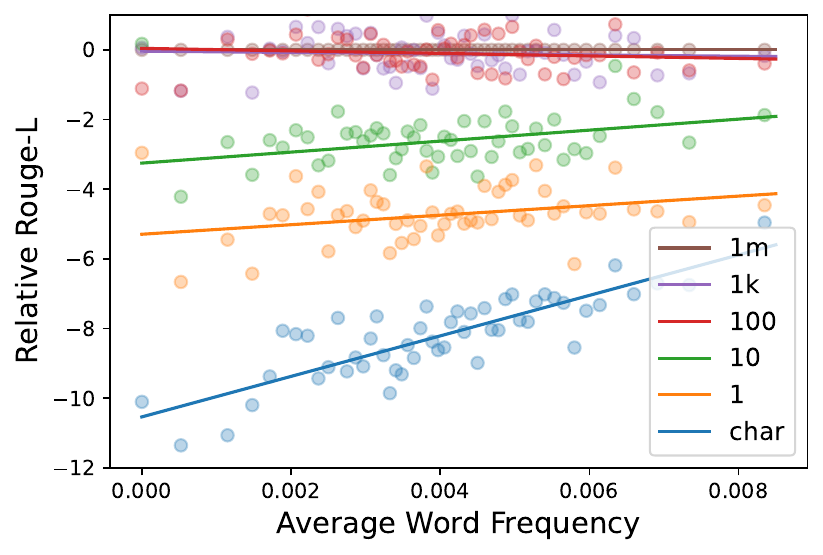}
    \caption{1B params model, XSum}
  \end{subfigure}
  
  \begin{subfigure}[b]{0.32\linewidth}
    \includegraphics[width=\linewidth]{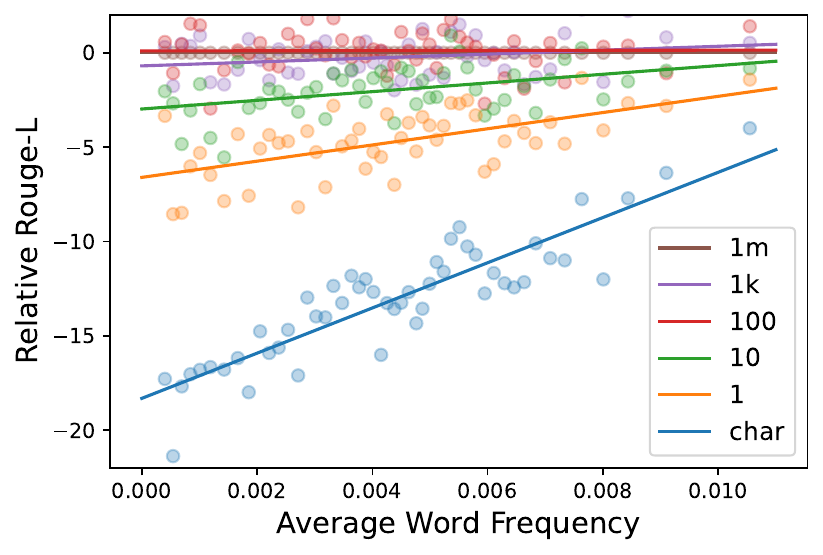}
    \caption{10m params model, QG-QA}
  \end{subfigure}
  \begin{subfigure}[b]{0.32\linewidth}
    \includegraphics[width=\linewidth]{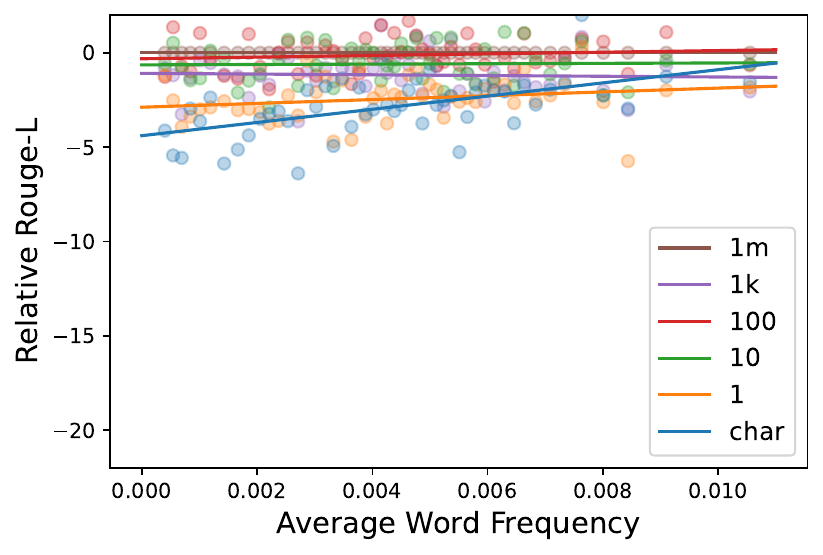}
    \caption{128m params model, QG-QA}
  \end{subfigure}
  \begin{subfigure}[b]{0.32\linewidth}
    \includegraphics[width=\linewidth]{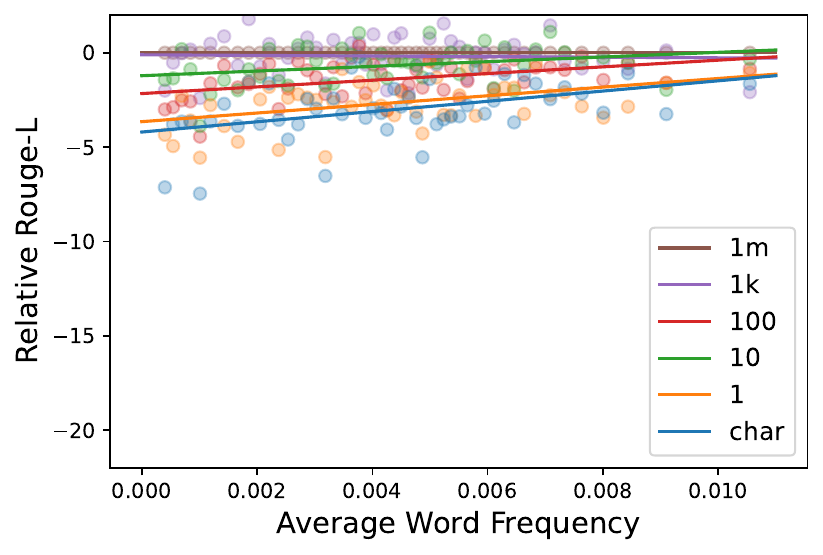}
    \caption{1B params model, QG-QA}
  \end{subfigure}
  \caption{Downstream success in Rouge-L relative to the \textsc{1m-doc} model plotted against the average frequency in each example. Trend lines were plotted based on the entire data, but for visibility reasons the scatter is based on averages over bins containing each 2\% of data.}
  \label{fig:dsVfreq}
\end{figure*}

\section{Analysis}

\paragraph{Tokenization of Frequent Words}
To better understand the source for discrepancy in compression between tokenizers, we plot in \autoref{fig:tokens_vs_freq} the number of tokens needed per word with respect to the word frequency (measured in number of appearances in a sample of 3 million unseen documents from mC4). %
We averaged the token-per-word ratio over all words whose occurrences are of the same order of magnitude and provided the number of words in each bin. A similar analysis was done for Turkish and it is shown in \autoref{fig:tokens_vs_freq_tur}. 

The figures show that the token-to-word ratio is extremely similar across tokenizers for words that are the most frequent. On the other hand, the different tokenizers diverge in token-to-word ratio when presented with rarer words, with less supported tokenizers being more sensitive to word frequency, compared to better supported tokenizers. It is worth noting that the same trend applies to the \textsc{char} tokenizer, for which the number of tokens per word is simply its length in characters. This should not be surprising due to the tendency of frequently used words to be shorter in accordance to Zipf's law of abbreviation \cite{zipf1949human}.

In addition, as predicted by Zipf's law \cite{estoup1912gammes, zipf1949human}, %
the number of frequent words over which the tokenizers agree is quite small, In terms of types, but they cover a large portion of the 3 million document sample over which the statistics were calculated. The English words that appear at least $10^6$ times in the sampled corpus, 162 in number, cover 47\% of the words in the corpus. On the other hand, in Turkish, due to the thicker tail of the Zipf's distribution of morphologically rich languages, only 71 words answer this criterion, covering 26\% of the corpus. %

We conclude that the discrepancy in compression ability, as evident in \autoref{tab:compression} stems mostly from the difference in the compression in less common words. This tail of less frequent words is consisted of the semantically interesting words, so it is likely that this gap in compression causes the gaps in model performance.

\begin{figure*}[t]
  \centering
  \begin{subfigure}[b]{0.3\linewidth}
    \includegraphics[width=\linewidth]{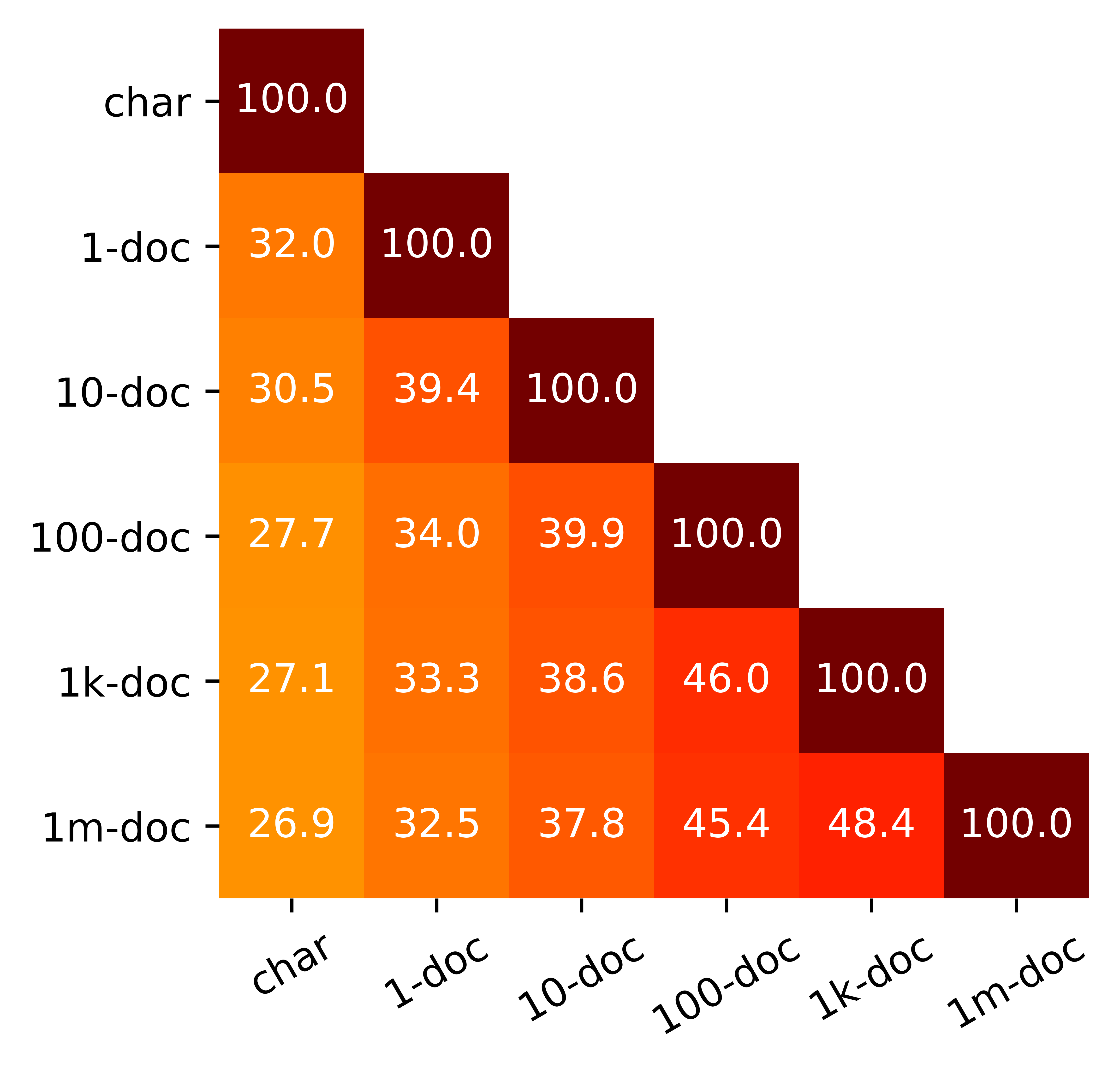}
    \caption{10m params model, XSum}
  \end{subfigure}
  \begin{subfigure}[b]{0.3\linewidth}
    \includegraphics[width=\linewidth]{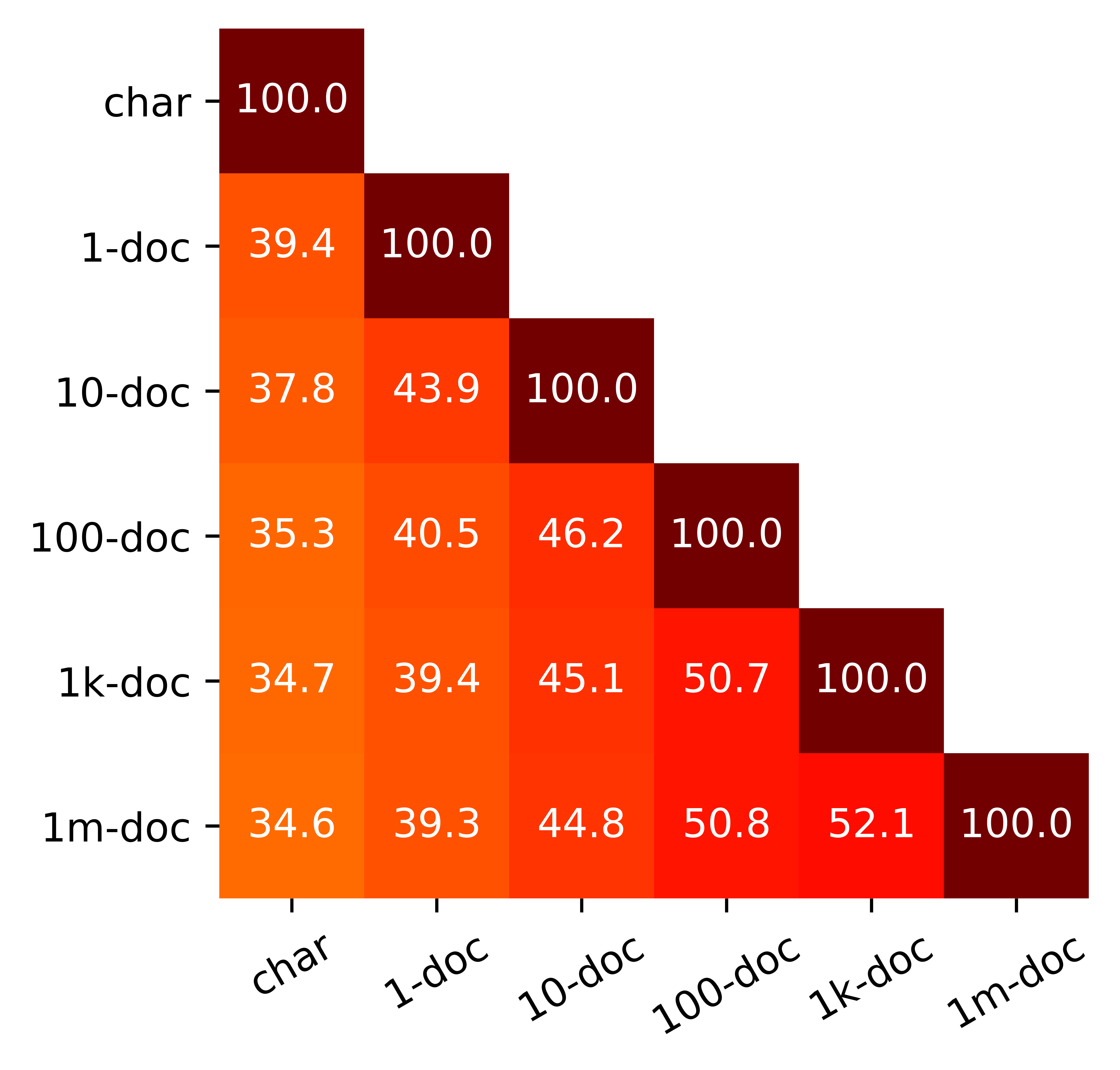}
    \caption{128m params model, XSum}
  \end{subfigure}
  \begin{subfigure}[b]{0.3\linewidth}
    \includegraphics[width=\linewidth]{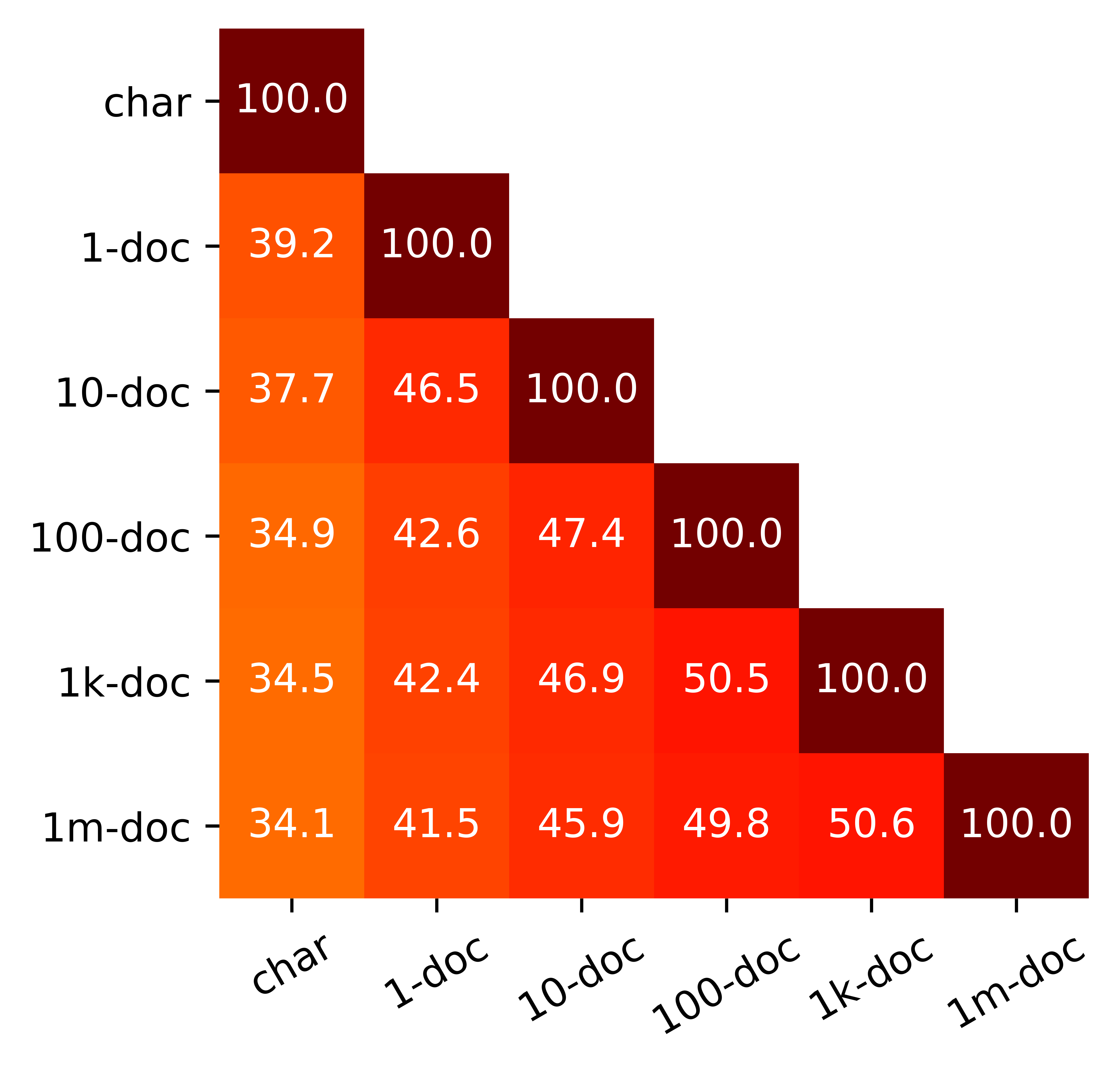}
    \caption{1B params model, XSum}
  \end{subfigure}
  
  \begin{subfigure}[b]{0.3\linewidth}
    \includegraphics[width=\linewidth]{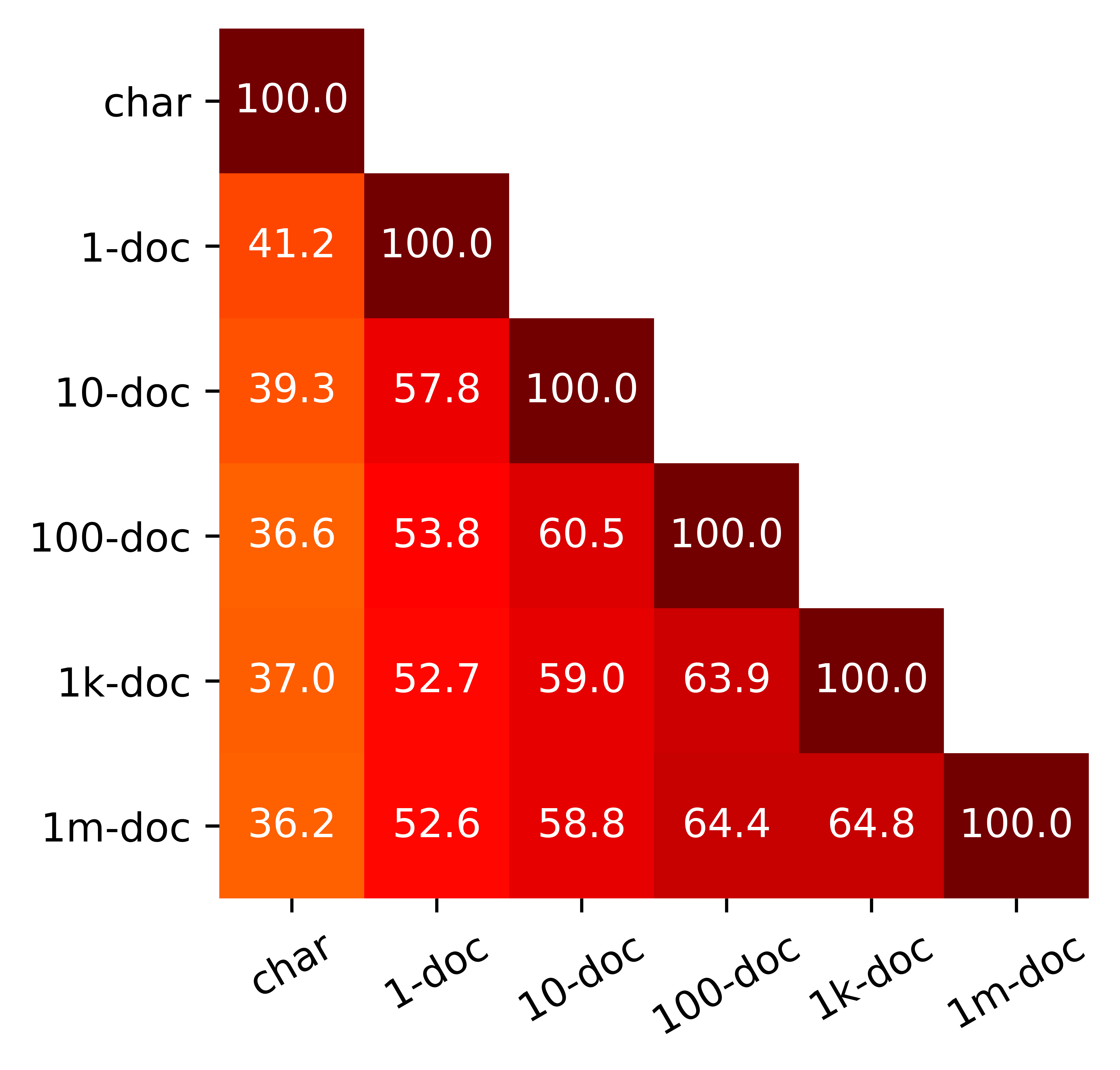}
    \caption{10m params model, QG-QA}
  \end{subfigure}
  \begin{subfigure}[b]{0.3\linewidth}
    \includegraphics[width=\linewidth]{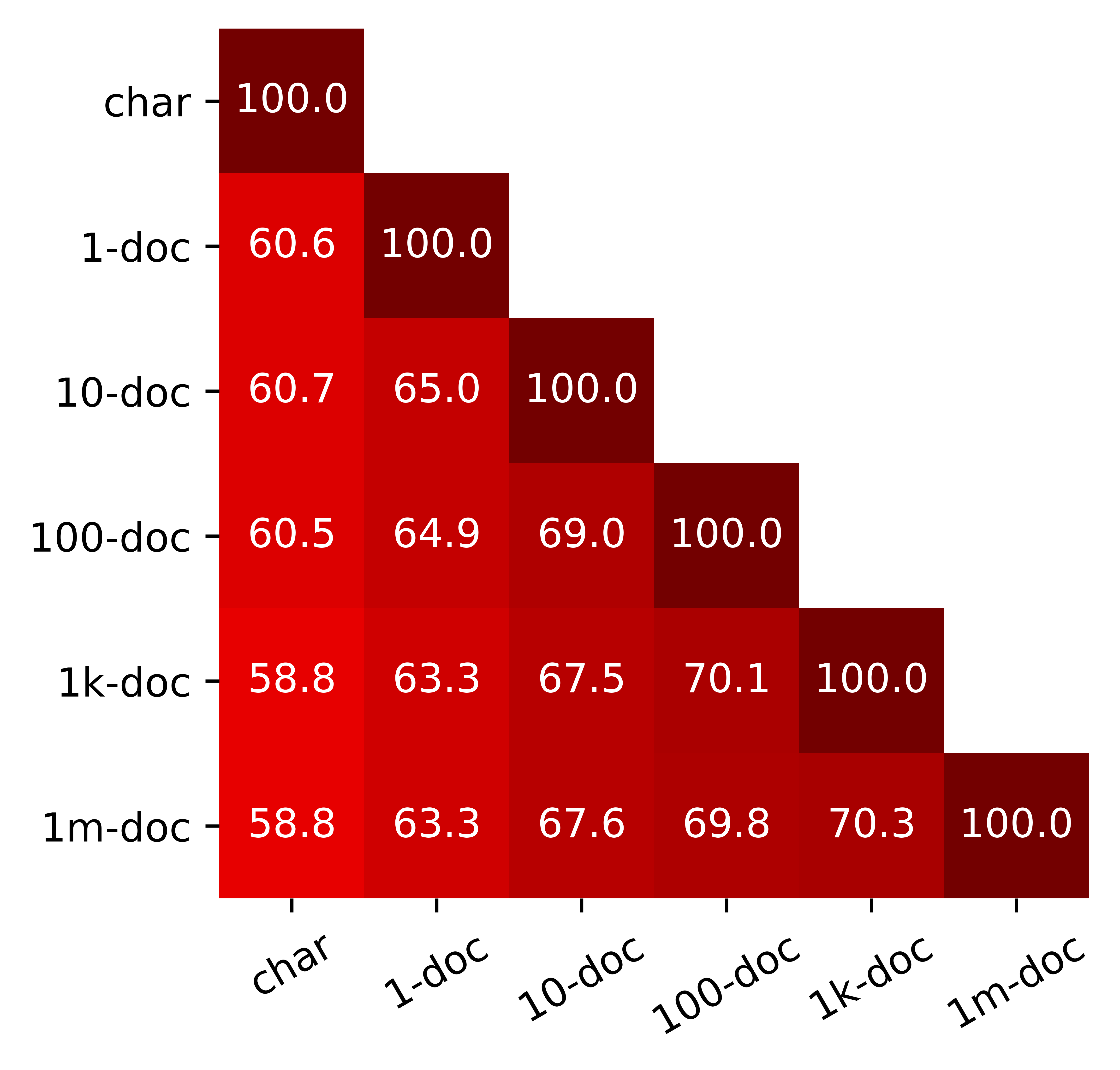}
    \caption{128m params model, QG-QA}
  \end{subfigure}
  \begin{subfigure}[b]{0.3\linewidth}
    \includegraphics[width=\linewidth]{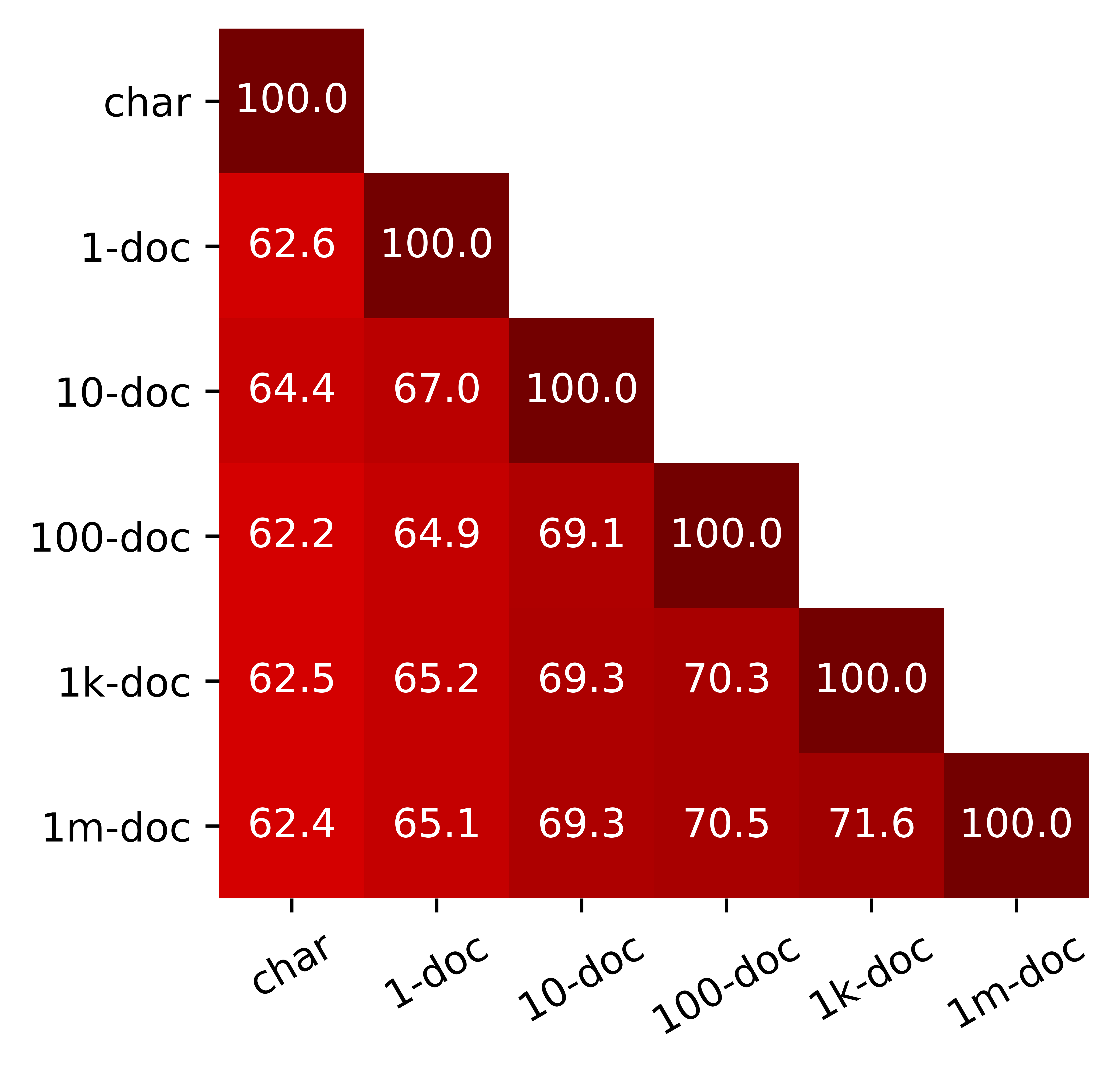}
    \caption{1B params model, QG-QA}
  \end{subfigure}
  \caption{Pair-wise rouge scores between outputs of the different models. Darker is higher Rouge-L scores and higher similarity between outputs. Models with similar number of supporting documents tend to output similar predictions.}
  \label{fig:grid}
\end{figure*}

\paragraph{Performance over Frequent Words} To complement the analysis above we also broke down the results of the generation tasks by average frequency of the words in the targeted output of each example. The results, plotted in \autoref{fig:dsVfreq}, include the difference in \mbox{Rouge-L} per example from the best \textsc{1m-doc} model per task and model size. it shows that the differences between differently tokenizing models are more pronounced over examples with rarer words. %

Together, this and the previous analysis shed some light on the reasons for the correlation found in our main result. We demonstrate that the differences in performance between the various models are indeed more pronounced in the presence of rarer words, which are exactly the words that the tokenizers compress differently. It is thus highly probable that word frequency is a major confounding factor the connects compression with downstream performance.

In addition, this analysis may point to the benefit of challenge sets, comprised of examples with rarer words, in the evaluation of tokenization.

\paragraph{Similarity between Tokenizers}
The results so far compared the output of each model to the target outputs, where we showed that models are performing better when equipped with better compressing tokenizers. In order to show that the models are also %
converging towards similar generations, we plotted, in \autoref{fig:grid}, the pair-wise overlap between the outputs of all models for the English generation tasks, measured in \mbox{Rouge-L} . %

The analysis shows that, for all tasks and model sizes, models with similarly supported tokenizers tend to output similar predictions, regardless of whether the predictions are similar to the gold targets.
It is also noticeable that, in accordance with our main results, the differences in the high-support region are less pronounced than those between the less supported tokenizers.

\section{Conclusions}

In this paper we demonstrated the importance of compression to tokenization as an intrinsic evaluation of tokenization quality that indicates the %
performance on extrinsic downstream tasks. We argued in favor of compression-driven tokenization from a theoretical perspective, %
since it may make the tokenizer act as a simple standalone language model, %
and we showed its correlation with downstream model success.

Our experiments point to generation tasks as better downstream evaluators of tokenization since their results are both more sensitive to the tokenizer and better correlate with tokenization quality as expressed in compression ability.

In terms of linguistic diversity, the similarity in the results and analyses across two very different languages, English and Turkish, points to our conclusions being %
independent of specific typological characteristics.
Yet, ample room is left for studying the effects of tokenization on more languages that are even more typologically diverse. Moreover, other intrinsic evaluators are still to be assessed even for the languages we did work with.

We conclude that tokenization matters, as poorly compressing tokenizers hinder the results of language models, and that investment in better compressing tokenizer has a potential of improving model performance while being relatively cheap in terms of compute. We therefore call for research to better understand the factors controlling the quality of the tokenization and its relation to overall success of the LLMs. %

\section{Limitations}

The main limitation of this paper has to do with the amount of resources allocated to this research. Pretraining our LLMs, especially the 1B parameter models, requires a lot of compute and repeating these experiments, in a slightly different setting or just in order to replicate their results, is an expensive process.

Another limitation has to do with the limited experiments on non-English languages. Although we executed several experiments on Turkish, the cost of pretraining models of up to 1B parameters prevented us from equating the treatment given to the two languages as well as adding experiments in other non-English languages. It is possible, even if somewhat unlikely, that running the full suite of experiments on Turkish would have resulted in different conclusions. A more reasonable possibility is that running experiments on more typologically diverse languages would yield different conclusions for these languages. We mitigated this risk by choosing a language that is extremely different from English.

\section*{Acknowledgements}

We thank Alon Jacovi and Uri Shaham for the helpful discussions and feedback.

\bibliography{anthology,custom}

\appendix

\section{Training Details}
\label{sec:appendix}

\paragraph{Tokenizers} were trained on the first documents in C4, for English, and mC4, for Turkish, as they are ordered in the Tensorflow Datasets repository. Therefore, the data for the less-supported tokenizers is contained in the data for the better supported ones.

In all cases we limited the vocabulary size to 32$k$, but in practice, for tokenizers supported by little data, the vocabulary size was lower than 32$k$, since the data did not contain a sufficient number of words and subwords. Specifically, for English, the vocabulary size of \textsc{100-doc} is 23$k$, of \textsc{10-doc} -- 3.7$k$, and of \textsc{1-doc} -- 1$k$. For Turkish the size of \textsc{10-doc} is 9.5$k$, and of \textsc{1-doc} -- 3.9$k$.

In the case of \textsc{char}, supported by no documents at all, the tokenizer simply breaks all the words into characters and replaces any foreign character with the \emph{unknown} sign.

\paragraph{Models} were trained using the T5X framework \cite{roberts2022t5x} on the span corruption task \cite{2020t5} for 200$k$ training steps, with a batch size of 512. Every training example was truncated to the maximal length of 1024 tokens.\footnote{The fixed example length in terms of tokens leads of course to differences in the amount of data seen by the models during training based on their tokenizers. We consider this as another boon of well-compressing tokenizers, since the computation budget is usually preset, as it is the case in our setting.}

The models were finetuned for 4$k$ steps and 20$k$ steps on the classification and generation tasks, respectively, with a batch size of 128. The decoder-only models were tasks with the generation of a gold output when used for classification tasks as well. For example, in the QQP task the outputs were assumed to be either \textit{duplicated} or \textit{no duplicated}, where any other output considered wrong. A manual inspection showed that models learned perfectly to output one of the desired targets.

\end{document}